\documentclass[sigconf]{acmart}
\AtBeginDocument{%
  }

\setcopyright{acmlicensed}
\copyrightyear{2018}
\acmYear{2018}
\acmDOI{XXXXXXX.XXXXXXX}
\acmConference[Conference acronym 'XX]{Make sure to enter the correct
  conference title from your rights confirmation email}{June 03--05,
  2018}{Woodstock, NY}
\acmISBN{978-1-4503-XXXX-X/2018/06}

\usepackage[ruled,vlined,linesnumbered]{algorithm2e}
\usepackage{booktabs}
\usepackage{multirow}
\usepackage{tabularx}
\usepackage{siunitx}
\usepackage{adjustbox,makecell}
\usepackage{pifont}
\usepackage{xcolor}
\usepackage{arydshln}
\usepackage{hyphenat}
\usepackage{amsmath,mathtools}
\usepackage{placeins}
\usepackage{cuted}     
\usepackage{capt-of}   
\newcommand{\cmark}{\ding{51}}
\newcommand{\impr}[1]{{\color{red}\scriptsize\,(#1)}}
\DeclareUnicodeCharacter{2081}{$_1$}
\DeclareUnicodeCharacter{2082}{$_2$}
\begin{document}

\title{STaR: Towards Effective and Stable Table Reasoning via Slow-Thinking Large Language Models}

\author{Huajian Zhang}
\email{zhjustc@mail.ustc.edu.cn}
\affiliation{%
  \institution{State Key Laboratory of Cognitive Intelligence, University of Science and Technology of China}
  \city{Hefei}\country{China}
}

\author{Mingyue Cheng}
\authornote{Corresponding author.}
\email{mycheng@ustc.edu.cn}
\affiliation{%
  \institution{State Key Laboratory of Cognitive Intelligence, University of Science and Technology of China}
  \city{Hefei}\country{China}
}

\author{Yucong Luo}
\email{prime666@mail.ustc.edu.cn}
\affiliation{%
  \institution{State Key Laboratory of Cognitive Intelligence, University of Science and Technology of China}
  \city{Hefei}\country{China}
}

\author{Xiaoyu Tao}
\email{txytiny@mail.ustc.edu.cn}
\affiliation{%
  \institution{State Key Laboratory of Cognitive Intelligence, University of Science and Technology of China}
  \city{Hefei}\country{China}
}

\renewcommand{\shortauthors}{Huajian Zhang, Mingyue Cheng, Yucong Luo and Xiaoyu Tao}

\begin{abstract}
Table reasoning with large language models (LLMs) plays a critical role in building intelligent systems capable of understanding and analyzing tabular data. Despite recent progress, existing methods still face key limitations: their reasoning processes lacks depth and explicit multi-step reasoning, often relying solely on implicit language model understanding. In addition, their reasoning processes suffer from instability, primarily caused by model uncertainty. In this work, we propose STaR, a novel slow-thinking model that can achieve effective and stable table reasoning. To enable effective multi-step reasoning, we design a two-stage training framework consisting of supervised fine-tuning (SFT) warm-up followed by reinforced fine-tuning (RFT). Specifically, in the SFT stage, we construct a high-quality dataset through automatic self-verification. In the RFT stage, we introduce a difficulty-aware reinforcement learning mechanism to further enhance reasoning capabilities. Furthermore, to improve reasoning stability, we introduce trajectory-level uncertainty quantification, which fuses token-level confidence with answer-level consistency, enabling the selection of better reasoning trajectories. Extensive experiments demonstrate that STaR-8B achieves state-of-the-art performance on in-domain benchmarks and exhibits strong generalization to out-of-domain datasets, highlighting its potential for enhancing both effectiveness and stability in table reasoning.\footnote{The code is publicly available at \url{https://github.com/zhjai/STaR}}
\end{abstract}


\keywords{Table Reasoning, Reinforced Fine-Tuning, Uncertainty Quantification}
\maketitle

\section{Introduction}
Tables represent one of the most prevalent structured data formats, serving as essential carriers of structured information across scientific publications, financial reports, and enterprise databases \cite{cheng2025survey,cheng2023formertime}. Over the past few years, researchers increasingly focused on mining tabular data. Among various table mining tasks, table reasoning is particularly fundamental, as it requires understanding complex table structures, integrating information across rows and columns, and performing multi-step logical and numerical reasoning to derive accurate answers \cite{zhang2025survey}.

Prior work on table reasoning is largely built on an execution grounded paradigm, where models translate a natural language question into an explicit, executable representation and run it over the table to obtain the answer \cite{cheng2025survey}. Representative lines include semantic parsing, which maps questions to logical forms executable on semi structured tables \cite{pasupat2015compositional}, and text to SQL, which synthesizes SQL queries conditioned on the question and schema for database execution \cite{yavuz2018takes,yu2018spider}. To handle more realistic settings that require combining evidence across multiple cells and aligning tables with accompanying text, researchers further developed structured pipelines such as multi-hop table reasoning \cite{chen2020hybridqa} and graph-based evidence composition \cite{zhu2023soargraph}. More recently, large language models (LLMs) have introduced a different family of approaches that rely primarily on pretrained language understanding. These methods typically serialize tables as text and adapt pretrained models to read and manipulate tabular structures \cite{zhang2024tablellama}. Some guide generation with structure-aware intermediate steps aligned with table operations \cite{jiang2023structgpt}, while others improve reliability via retrieval or program-based computation \cite{chen2024tablerag,mao2024potable}. Overall, most LLMs-based table reasoning approaches mainly use the model's language understanding to process tables and produce answers.

Despite these advances, existing LLMs-based methods remain constrained by two critical limitations \cite{zhang2025survey,liu2024rethinking}. First, they lack slow-thinking capabilities, typically failing to perform explicit, multi-step reasoning required for complex table reasoning. Second, they exhibit reasoning instability, often leading to unreliable outcomes in decision-making scenarios. Moreover, practical deployment often demands lightweight, privacy-preserving models rather than extremely large LLMs. Motivated by above analysis, we aim to construct a lightweight model that embodies both explicit slow-thinking reasoning and high stability. However, achieving this requires overcoming two major challenges that have been largely overlooked in prior research: (i) how to effectively train slow-thinking model specific to table reasoning, and (ii) how to stabilize the reasoning trajectories to ensure consistent generation.

To address these challenges, we propose STaR, an effective and stable slow-thinking model for LLM-based table reasoning. STaR improves training effectiveness through a two-stage training framework, consisting of supervised fine-tuning (SFT) warm-up and reinforced fine-tuning (RFT). In the SFT stage, we construct a high-quality slow-thinking dataset via self-verification to enable explicit, step-by-step reasoning. In the RFT stage, we introduce a  difficulty-aware reinforcement learning (RL) mechanism to further strengthen multi-step reasoning on harder queries. Furthermore, to improve reasoning stability at inference time, we propose trajectory-level uncertainty quantification, which fuses token-level confidence with answer-level consistency to select better reasoning trajectories among multiple candidates. Extensive experiments show that STaR-8B achieves state-of-the-art results on in-domain benchmarks and generalizes well to out-of-domain datasets, demonstrating its potential to improve both effectiveness and reliability in table reasoning. Our main contributions are:

\begin{itemize}
    \item We propose STaR, a slow-thinking table reasoning model that integrates explicit multi-step reasoning with uncertainty quantification to achieve both effective and stable reasoning.
    \item We construct a self-verified slow-thinking dataset for SFT warm-up, a difficulty-aware RL mechanism to enhance reasoning capability, and a trajectory-level uncertainty quantification method that fuses token-level confidence with answer-level consistency to improve stability.
    \item We conduct extensive experiments on multiple benchmarks, demonstrating that STaR-8B achieves state-of-the-art accuracy with improved stability and strong generalization to out-of-domain datasets.
\end{itemize}

\section{Related Work}

\begin{figure*}[t]
  \centering
  \includegraphics[width=\textwidth]{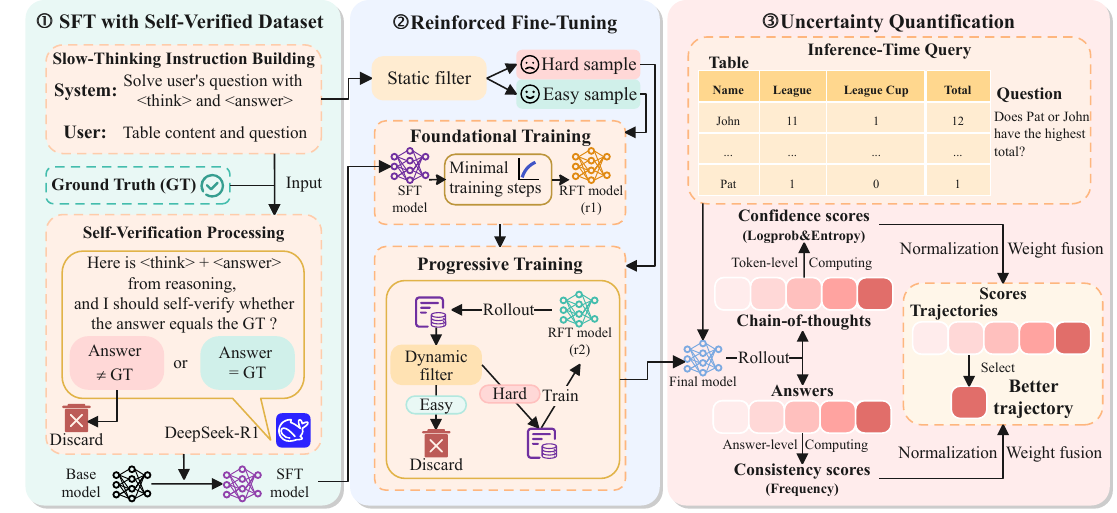}
  \caption{Overview of the STaR framework, including SFT warm-up with self-verified dataset, reinforced fine-tuning, and trajectory-level uncertainty quantification.}
  \label{fig:table_r2_framework}
\end{figure*}

This section reviews three key areas: table reasoning, LLMs-based reasoning, and uncertainty quantification.

\subsection{Table Reasoning}
Table reasoning has evolved from early neural semantic parsers to modern LLM-based methods. Benchmarks such as WikiTableQuestions \cite{pasupat2015compositional}, TabFact \cite{chen2019tabfact}, and FinQA \cite{chen2021finqa} drive progress and evaluation in this area. Recent progress has concentrated on several key areas: Chain-of-Table \cite{wang2024chain} introduces dynamic table transformations for multi-step reasoning; TaPERA \cite{zhao2024tapera} improves accuracy in long-form question answering; and tool-enhanced frameworks such as TabSQLify \cite{nahid2024tabsqlify}, TableRAG \cite{chen2024tablerag}, TableMind \cite{jiang2025tablemind}, and text-to-SQL methods \cite{lei2025reasoning,dong2025reasoning} utilize external tools to enhance their capabilities. Additional areas include multimodal understanding \cite{su2024tablegpt2} and data augmentation \cite{zhang2024tablellm}. Reinforcement learning is also increasingly applied, particularly in methods focused on inference-time scaling \cite{yang2025table}, region-based optimization \cite{wu2025table}, and program-based reasoning \cite{jin2025table}. Despite these advances, existing methods often lack explicit multi-step reasoning processes. STaR addresses this limitation through a two-stage training framework that combines SFT warm-up with RFT, enabling more effective table reasoning with explicit multi-step reasoning.

\subsection{LLMs-Based Reasoning}
Enhancing the reasoning capabilities of LLMs become a central research focus. Slow-thinking emerges as a leading paradigm that enables explicit multi-step reasoning through extended thinking processes \cite{cheng2025can,luo2025time}. Frontier models such as OpenAI GPT-5 \cite{leon2025gpt}, DeepSeek-R1 \cite{guo2025deepseek}, Google Gemini 2.5 \cite{comanici2025gemini}, Kimi K2 \cite{team2025kimi}, and Qwen3 \cite{yang2025qwen3} often incorporate reinforcement-learning-based post-training to enhance reasoning performance. Training efficiency is improved by algorithms like GRPO \cite{shao2024deepseekmath}, which eliminates the critic model to reduce memory usage, and its successor DAPO \cite{yu2025dapo}, which further cuts down training steps. The RLVR approach \cite{lambert2024tulu} enables scalable training by using automated verification, removing the need for expensive human annotation. A notable trend is the shift to process-level supervision, which is often more effective than rewards that consider the final answer. For example, PRMs \cite{lightman2023let} use step-wise feedback, while recent innovations like ThinkPRM \cite{khalifa2025process} and PAVs \cite{setlur2024rewarding} improve data efficiency and progress measurement. Further gains are achieved through advanced curriculum strategies that provide theoretical guarantees or use bandit-based selection \cite{parashar2025curriculum, chen2025self}, and through test-time compute scaling, which allows smaller models to outperform much larger ones \cite{snell2024scaling}. Despite their success in mathematical and coding domains, these advances are not systematically adapted to table reasoning. STaR bridges this gap by extending slow-thinking mechanisms to the tabular domain, enabling explicit multi-step reasoning over structured data.

\subsection{Uncertainty Quantification}
Uncertainty quantification is important for the reliability of LLMs. Token-level approaches, including entropy-based methods \cite{kuhn2023semantic} and confidence scores \cite{lin2022teaching}, provide accurate estimates, while semantic entropy \cite{farquhar2024detecting} aggregates semantically similar outputs before calculating uncertainty. Self-consistency decoding \cite{wang2022self} samples multiple reasoning paths via majority voting, improving GSM8K by +17.9\% over greedy decoding. Tree-of-Thoughts \cite{yao2023tree} and ReAct \cite{yao2023react} integrate systematic path evaluation. Recent advances include confidence-informed self-consistency reducing samples by 40\% \cite{taubenfeld2025confidence}, universal self-consistency for free-form generation \cite{chen2023universal}, and Kernel Language Entropy \cite{nikitin2024kernel}. Conformal prediction \cite{quach2023conformal} provides distribution-free guarantees, while calibration techniques like THERMOMETER \cite{shen2024thermometer} and APRICOT \cite{ulmer2024calibrating} adapt temperature scaling. However, most methods operate at the token or answer level, overlooking the structured nature of reasoning trajectories. STaR addresses this by introducing trajectory-level uncertainty quantification that fuses token-level confidence with answer-level consistency to select better reasoning trajectories.

Building upon these advances, STaR addresses key limitations in table reasoning through two core contributions. For effectiveness, we introduce a two-stage training framework that combines SFT warm-up with difficulty-aware RFT, progressively enhancing reasoning capabilities. For stability, STaR performs trajectory-level uncertainty quantification during inference by fusing token-level confidence with answer-level consistency, enabling the selection of better reasoning trajectories.

\section{Methodology}

This section presents STaR, which integrates three key components: SFT warm-up with self-verified dataset, reinforced fine-tuning, and uncertainty quantification.

\subsection{Framework Overview}
As illustrated in Figure \ref{fig:table_r2_framework}, STaR integrates three key components: SFT warm-up with self-verified dataset, reinforced fine-tuning, and uncertainty quantification. The framework employs a two-stage training process consisting of SFT warm-up followed by RFT. In the SFT stage, we construct high-quality training data through structured prompts with self-verification mechanism. In the RFT stage, we apply difficulty-aware RL with foundational training on easy samples followed by progressive training on hard samples through iterative refinement with dynamic sample filtering, progressively expanding the model's capability boundary. During inference, STaR generates multiple reasoning trajectories and quantifies their uncertainty by fusing token-level confidence with answer-level consistency, enabling it to select the trajectory with the highest composite score. This trajectory-level uncertainty quantification enhances both accuracy and stability in table reasoning, especially for multi-step reasoning tasks.

\subsection{SFT with Self-Verified Dataset}
\label{sec:sft}

\subsubsection{Slow-Thinking Data Construction}
To establish foundational table reasoning capabilities, we construct a training dataset from WikiTableQuestions, HiTab, and FinQA. We serialize tables into markdown format to preserve structural information while maintaining readability, then prompt DeepSeek-R1 with the table, question, and ground-truth answer to generate complete reasoning trajectories. The generated demonstrations follow a structured format: \texttt{<think>} reasoning process \texttt{</think>} followed by \texttt{<answer>} final answer \texttt{</answer>}, where the thinking section contains detailed chain-of-thought (CoT) reasoning with intermediate calculations and the answer section provides structured JSON output in the required format. This slow-thinking format encourages models to engage in explicit multi-step reasoning before reaching conclusions, ensuring consistent SFT initialization and reducing mismatch between SFT and RFT prompts during training. The ground-truth answer is used only during dataset construction. Complete prompt templates are provided in Appendix \ref{sec:prompt-templates}.

\subsubsection{Self-Verification Mechanism}
In data construction, we employ a self-verification mechanism, where the model compares its generated answer against the ground truth and automatically filters out samples with inconsistent outputs. This step ensures alignment between reasoning trajectories and final answers, and removes invalid JSON outputs, thereby significantly improving data quality and reliability. Through this dataset with self\hyp{}verified quality control, models acquire diverse table reasoning abilities including systematic information extraction, multi-step reasoning with explicit intermediate computations, self\hyp{}reflection, and structured answer generation, providing a strong foundation for the subsequent RFT.

\subsection{Reinforced Fine-Tuning}
\subsubsection{GRPO Objective} Following recent advances in long CoT training, we adopt the GRPO \cite{shao2024deepseekmath} variant introduced in DAPO \cite{yu2025dapo}, which removes the KL divergence penalty and employs asymmetric clipping bounds. We refer to this variant as enhanced GRPO. The KL removal allows the model to diverge from its initial distribution when discovering complex reasoning strategies required for table reasoning, while the asymmetric clipping strategy encourages exploration of alternative reasoning trajectories during training. These modifications are important for table reasoning, where diverse problem-solving approaches across varying table complexities require flexible policy adaptation. Our GRPO objective becomes:
\begin{equation}
    \begin{aligned}
    \mathcal{J}_{\mathrm{GRPO}}(\theta) & =\mathbb{E}_{(q, a) \sim \mathcal{D},\{o_{i}\}_{i=1}^{G} \sim \pi_{\theta_{\text{old}}}(\cdot \mid q)} \\
    & \Bigl[\frac{1}{\sum_{i=1}^{G} |o_i|} \sum_{i=1}^{G} \sum_{t=1}^{|o_i|} \min\left(r_{i,t}(\theta)\hat{A}_{i,t},\right. \\
    & \left.\operatorname{clip}\left(r_{i,t}(\theta), 1-\varepsilon_{\text{low}}, 1+\varepsilon_{\text{high}}\right)\hat{A}_{i,t}\right)\Bigr],
    \end{aligned}
\end{equation}
\noindent where $G$ denotes the number of sampled trajectories per query, $|o_i|$ is the token length of trajectory $o_i$, $r_{i,t}(\theta)=\frac{\pi_{\theta}(o_{i,t}\mid q,o_{i,<t})}{\pi_{\theta_{\text{old}}}(o_{i,t}\mid q,o_{i,<t})}$ is the token-level policy ratio, and $\hat{A}_{i,t}$ is the corresponding advantage. We use asymmetric clipping bounds with $\varepsilon_{\text{high}}>\varepsilon_{\text{low}}$ (e.g., $\varepsilon_{\text{high}}$=0.28, $\varepsilon_{\text{low}}$=0.2) to allow larger updates that reinforce beneficial reasoning abilities while constraining harmful ones.

\subsubsection{Difficulty-Aware Training} As illustrated in Figure \ref{fig:two-stage-grpo} (Appendix \ref{sec:difficulty-aware-rl-pipeline}), our RFT training strategy employs difficulty-aware RL by separating easy and hard reasoning tasks. We split training samples into an easy subset $S_{\text{easy}}$ and a hard subset $S_{\text{hard}}$ based on pass@k difficulty assessment. Foundational training uses $S_{\text{easy}}$ to obtain an intermediate policy $\pi_{\text{interim}}$; progressive training initializes from it, trains on $S_{\text{hard}}$, and outputs the final policy $\pi_{\text{final}}$.

During foundational training, we train $\pi_{\text{interim}}$ on $S_{\text{easy}}$ with high learning rates (e.g., $1\times 10^{-5}$), enabling the model to quickly achieve about 80\% performance within minimal training steps. This establishes foundational reasoning abilities without the additional complexity of processing hard samples. A uniform training approach would require extensive time for the model to learn and adaptively filter these easy samples at lower learning rates, which would reduce the efficiency of training.

During progressive training, we start from $\pi_{\text{interim}}$, train on $S_{\text{hard}}$ with lower learning rates (e.g., $1\times 10^{-6}$), and output $\pi_{\text{final}}$. It uses dynamic sample filtering based on pass@k$_2$ evaluation. The filtering mechanism operates as follows: To avoid wasting updates on solved samples, samples with pass@k$_2$=1.0 are permanently excluded; samples with pass@k$_2$<1.0 are placed into a review pool for periodic reevaluation, and only samples with pass@k$_2$<0.8 receive active GRPO updates. This dynamic filtering ensures that computational resources focus on difficult reasoning problems:
\begin{equation}
     \mathcal{J}_{\mathrm{GRPO}} = {\mathbb{E}}_{s \in {\mathcal{S}}_{\text{active }}} \left[  {{\mathcal{L}}_{\mathrm{GRPO}}\left( s\right) }\right]  \;  s.t.  \;0 \le pass@k_{2}\left( s\right) < {0.8},
\end{equation}
\noindent where \(\mathcal{S}_{\text{active}}\) represents the actively trained sample subset. This strategy ensures that training resources are concentrated on samples at the model's appropriate difficulty level, avoiding computational waste on overly simple examples while maintaining previously learned capabilities.

Compared to uniform training that processes all samples together, this difficulty-aware approach is more efficient: foundational training quickly masters easy samples with higher learning rates, while progressive training concentrates computational resources on challenging problems with lower learning rates.

\subsubsection{Reward Function Design} We employ a composite reward function that combines three components \(R_{\text{format}}\), \(R_{\text{partial}}\), and \(R_{\text{complete}}\). Each component is binary, and we weight them by 0.2, 0.3, and 0.5, respectively. Format correctness \(R_{\text{format}}\) checks whether the \texttt{<think>} and \texttt{<answer>} tags are correctly aligned and whether the answer section is valid JSON, since incorrect outputs cannot be parsed or evaluated. Partial correctness uses \(R_{\text{partial}} = \mathbb{I}(\hat{y} \cap y \neq \emptyset)\) to reward predictions that overlap with any ground truth element, providing intermediate learning signals to reduce reward sparsity. Complete correctness \(R_{\text{complete}}\) provides the primary learning signal for exact matches with the ground truth. The final reward is computed as \(R = 0.2 \cdot R_{\text{format}} + 0.3 \cdot R_{\text{partial}} + 0.5 \cdot R_{\text{complete}}\), balancing format requirements and accuracy. It encourages models to generate both parseable and correct outputs.

\subsection{Uncertainty Quantification}

\subsubsection{Token-Level Confidence Metrics} We estimate token-level confidence for each reasoning trajectory from the model's internal probability distributions. For a generated trajectory \(r\) consisting of tokens \(\{t_1, t_2, ..., t_n\}\), we compute the average log-probability and normalized entropy:
\begin{equation}
    \text{logprob}(r) = \frac{1}{n} \sum_{i=1}^{n} \log p(t_i \mid t_{<i}, q),
\end{equation}
\begin{equation}
    \begin{split}
    \text{entropy}(r) = -\frac{1}{n\log|\mathcal{V}|} \sum_{i=1}^{n} \sum_{v \in \mathcal{V}} \\
    \qquad p(v \mid t_{<i}, q) \log p(v \mid t_{<i}, q),
    \end{split}
\end{equation} 
\noindent where \(r\) denotes a generated reasoning trajectory for the query \(q\), \(\mathcal{V}\) denotes the vocabulary, and \(p(\cdot \mid t_{<i}, q)\) is the distribution over the next token given the previous tokens. The \(\log|\mathcal{V}|\) term normalizes entropy to \([0,1]\), where higher log-probability and lower entropy indicate lower uncertainty (higher confidence).

\begin{algorithm}[t]
\caption{Consistency-Confidence Fusion Selection}
\label{alg:ccfs-2e}
\KwIn{$R=\{r_1,\dots,r_n\}$, $L$ (log-probability), $E$ (entropy)}
\KwOut{Selected best response}
\SetKwFunction{Extract}{extract\_answer}
\SetKwFunction{IsValid}{isValid}
\SetKw{KwNormalize}{Normalize}

\texttt{answer\_groups} $\gets \varnothing$\;
\ForEach{$(r_i, l_i, e_i) \in (R,L,E)$}{
  $a \gets \Extract(r_i)$ \tcp*[r]{answer}
  \If{\IsValid{$a$}}{
    $c \gets \exp(l_i)\times (1-e_i)$ \tcp*[r]{confidence}
    $\texttt{answer\_groups}[a] \gets \texttt{answer\_groups}[a] \cup \{(\text{response}=r_i,\ \text{confidence}=c)\}$\;
  }
}
\ForEach{$a \in \text{keys}(\texttt{answer\_groups})$}{
  $\texttt{consistency}[a] \gets |\texttt{answer\_groups}[a]|$\;
  $\texttt{avg\_conf}[a] \gets \mathrm{mean}\{c:(\_,c)\in\texttt{answer\_groups}[a]\}$\;
  $\texttt{max\_conf}[a] \gets \max\{c:(\_,c)\in\texttt{answer\_groups}[a]\}$\;
}
\KwNormalize $\texttt{consistency}$, $\texttt{avg\_conf}$, $\texttt{max\_conf}$ by their maxima across groups\;
\ForEach{$a \in \text{keys}(\texttt{answer\_groups})$}{
  $\texttt{final\_score}[a] \gets w_{\text{cons}}\,\widehat{\texttt{consistency}}[a] + w_{\text{avg}}\,\widehat{\texttt{avg\_conf}}[a] + w_{\text{max}}\,\widehat{\texttt{max\_conf}}[a]$\;
}
\Return response in $\texttt{answer\_groups}\bigl[\arg\max_a \texttt{final\_score}[a]\bigr]$ with the highest confidence\;
\end{algorithm}

\subsubsection{Consistency-Confidence Fusion Selection} Our trajectory\hyp{}level uncertainty quantification integrates token-level confidence with answer-level consistency, addressing the limitations of using either signal alone. Using only answer frequency can miss rare but correct solutions when most sampled trajectories yield plausible but incorrect answers. Token-level confidence alone can select overconfident but incorrect candidates. We group sampled trajectories by their final answer. We compute a score \(S(a)\) for each answer group. We then return the trajectory with the highest confidence from the group with the highest score, as shown in Algorithm \ref{alg:ccfs-2e}.

The mathematical formulation of our fusion strategy combines three normalized components:
\begin{equation}
    S(a) = w_{\text{cons}} \cdot \frac{|G_a|}{|G_{\text{max}}|} + w_{\text{avg}} \cdot \frac{\bar{C}_a}{C_{\text{max}}} + w_{\text{max}} \cdot \frac{C_a^{\text{max}}}{C_{\text{max}}^{\text{global}}},
\end{equation}
\noindent where \(G_a\) is the set of trajectories whose final answer is \(a\); \(|G_a|\) is the answer frequency and \(|G_{\text{max}}|=\max_a |G_a|\). \(C\) denotes a confidence score from log-probability and entropy (Algorithm \ref{alg:ccfs-2e}). \(\bar{C}_a\) and \(C_a^{\text{max}}\) are the average and maximum confidence in group \(a\); we normalize by \(C_{\text{max}}=\max_a \bar{C}_a\) and \(C_{\text{max}}^{\text{global}}=\max_a C_a^{\text{max}}\).

\subsubsection{Weight Calibration and Analysis} We used grid search optimization \cite{hsu2003practical} on the training splits (Appendix \ref{sec:dataset-statistics}) to select weights for our trajectory-level uncertainty quantification, and kept them fixed for evaluation on the test splits. We set \(w_{\text{cons}}=0.25\), \(w_{\text{avg}}=0.2\), and \(w_{\text{max}}=0.55\).

The strong emphasis on maximum confidence (\(w_{\text{max}} = 0.55\)) is important because correct solutions may appear in only a few sampled trajectories while most produce plausible but incorrect answers. Maximum confidence can identify these high-quality trajectories that consistency metrics would miss. Furthermore, this trajectory-level metric is more stable than simple averaging, as it reduces the influence of noisy trajectories. This helps convert pass@k potential into improved pass@1 performance.

\section{Experiments}
This section covers experimental settings, performance evaluation, ablation studies on key components, and case studies. Additional experiments appear in Appendix \ref{sec:enhance-grpo-comparation}.

\subsection{Experimental Settings}
\subsubsection{Datasets} We conduct experiments on more than 30,000 table reasoning samples with various benchmarks. The training data is sourced from WikiTableQuestions (WTQ) \cite{pasupat2015compositional} for table QA, HiTab \cite{cheng2021hitab} for hierarchical table reasoning, and FinQA \cite{chen2021finqa} for numerical reasoning over financial tables. Evaluation includes both in-domain testing on the WTQ, HiTab, and FinQA test sets and out-of-domain generalization on TabMWP \cite{lu2022dynamic} for mathematical reasoning and TabFact \cite{chen2019tabfact} for fact verification. TabFact is a table-based fact verification task with binary entailment judgments; unlike question answering, it tests statement verification against tabular evidence and thus complements our evaluation of generalization and reasoning stability. Quality control is performed as described in Section \ref{sec:sft}. Detailed dataset statistics are provided in Appendix \ref{sec:dataset-statistics}.

\subsubsection{Baselines} We evaluate STaR against a wide range of state-of-the-art baselines across various categories to ensure thorough comparison of different reasoning approaches. In Table \ref{tab:overall-performance}, the CoT column indicates whether explicit step-by-step reasoning is enabled during inference, consistent with the slow-thinking setting. Our baselines include recent closed-source models with and without slow-thinking capabilities, reflecting recent progress in commercial LLMs development. We also compare our work to open-source models, including general LLMs and those that have been specifically improved with slow-thinking mechanisms. These models vary in scale from 0.6B to over 120B parameters. Additionally, we compare with specialized table reasoning systems that employ unique architectures or training strategies for tabular data. These include models fine-tuned for table tasks and those trained with reinforcement learning on verifiable table reasoning rewards. This diverse baseline selection allows us to evaluate STaR's performance across models of different scales, designs, and training methodologies.

\subsubsection{Implementation Details} We employ Qwen3-0.6B and Qwen3-8B as base models in STaR and train on 4×NVIDIA A800 80GB GPUs. The SFT stage employs a batch size of 256 with a learning rate of \(1\times 10^{-5}\) for 3 epochs. We use the SFT-tuned Qwen3-0.6B model to split the training data for the RFT training, based on pass@k$_1$ with $k_1=32$ and a threshold of 0.6. Foundational training establishes basic reasoning capabilities using batch size 512, 5 rollouts per sample, and learning rate \(1\times 10^{-5}\) over approximately 10,000 easy samples for 20 update steps, which roughly covers the easy subset once given batch size 512. Progressive training focuses on hard samples with batch size 256, 8 rollouts, and learning rate \(1\times 10^{-6}\) with weight decay 0.01 for 168 steps, resulting in a total of 188 steps, which aligns with the number of steps used in uniform GRPO baselines for a fair comparison. For decoding, both foundational and progressive training use temperature 1.0 with a maximum response length of 4096 tokens. During inference, we generate 8 rollouts per query with temperature 0.6 and top-$p$=0.95 (top-$k$ disabled), with maximum prompt and response lengths of 8192 and 4096 tokens. We generate rollouts in batches (1024 examples) and record log-probabilities and entropies for uncertainty quantification and trajectory selection.

\subsection{Table Reasoning Performance}

\begin{table*}[t]
  \centering
  \caption{Overall performance measured by exact match (EM) on in-domain (WTQ, HiTab, FinQA) and out-of-domain (TabMWP, TabFact) datasets. Improv. shows gains of STaR-8B over each baseline; \textcolor{red}{\scriptsize(+gain)} shows improvement over base models (Qwen3-0.6B/Qwen3-8B). Full performance in Appendix \ref{sec:full-overall-performance}}
  \label{tab:overall-performance}
  \footnotesize
  \setlength{\tabcolsep}{1pt}
  \renewcommand{\arraystretch}{1.10}
  \begin{tabular*}{\textwidth}{@{\extracolsep{\fill}} l l c cc cc cc cc cc @{}}
    \toprule
    \multicolumn{3}{l}{} &
      \multicolumn{6}{c}{\textbf{In-domain Performance}} &
      \multicolumn{4}{c}{\textbf{Out-of-domain Performance}} \\
    \cmidrule(lr){4-9}\cmidrule(lr){10-13}
    \textbf{Type} & \textbf{Baseline} & \textbf{CoT} &
      \multicolumn{2}{c}{\textbf{WTQ}} &
      \multicolumn{2}{c}{\textbf{HiTab}} &
      \multicolumn{2}{c}{\textbf{FinQA}} &
      \multicolumn{2}{c}{\textbf{TabMWP}} &
      \multicolumn{2}{c}{\textbf{TabFact}} \\
    \cmidrule(lr){4-5}\cmidrule(lr){6-7}\cmidrule(lr){8-9}\cmidrule(lr){10-11}\cmidrule(lr){12-13}
     & &  & EM (\%) & Improv. (\%) & EM (\%) & Improv. (\%) & EM (\%) & Improv. (\%) & EM (\%) & Improv. (\%) & EM (\%) & Improv. (\%) \\
    \midrule
    \multirow{4}{*}{Closed-Source}
      & GPT-4.1                 &        & 58.42 & 33.85 & 60.64 & 32.32 &  6.36 & 49.70 & 37.73 & 59.63 & 84.80 & 7.25 \\
      & GPT-5                   & \cmark & 90.10 &  2.17 & 43.96 & 49.00 & 29.21 & 26.85 & 55.18 & 42.18 & 91.20 & 0.85 \\
      & Gemini-2.0-flash        &        & 71.81 & 20.46 & 76.95 & 16.01 & 19.35 & 36.71 & 61.96 & 35.40 & 80.25 & 11.80 \\
      & Gemini-2.5-flash        & \cmark & 82.12 & 10.15 & 52.11 & 40.85 & 15.69 & 40.37 & 62.21 & 35.15 & 90.35 & 1.70 \\
    \midrule
    \multirow{2}{*}{Open-Source}
      & Qwen2.5-72B-Instruct    &        & 67.56 & 24.71 & 71.46 & 21.50 & 29.47 & 26.59 & 74.33 & 23.03 & 77.90 & 14.15 \\
      & DeepSeek-R1             & \cmark & 84.07 &  8.20 & 65.68 & 27.28 & 24.93 & 31.13 & 71.06 & 26.30 & 89.50 &  2.55 \\
    \midrule
    \multirow{3}{*}{\makecell[l]{Table-Reasoning\\Specific}}
      & TableGPT2-7B            &        & 47.60 & 44.67 & 63.11 & 29.85 & 15.54 & 40.52 & 53.71 & 43.65 & 21.16 & 70.89 \\
      & Table-R1-Zero-7B  & \cmark & 82.04 & 10.23 & 86.63 &  6.33 & 19.23 & 36.83 & 71.50 & 25.86 & 56.52 & 35.53 \\
      & Table-R1-Zero-8B  & \cmark & 83.09 &  9.18 & 88.61 &  4.35 & 14.46 & 41.60 & 43.78 & 53.58 & 87.63 &  4.42 \\

    \midrule
    \multirow{2}{*}{Our Base Models}
      & Qwen3-0.6B              & \cmark & 36.06 & 56.21 & 34.26 & 58.70 & 10.18 & 45.88 & 53.90 & 43.46 & 43.39 & 48.66 \\
      & Qwen3-8B                & \cmark & 83.29 & 8.98  & 70.04 & 22.92 & 26.63 & 29.43 & 64.76 & 32.60 & 90.22 &  1.83 \\
    \midrule
    \multirow{2}{*}{Our Models}
      & \textbf{STaR-0.6B} & \cmark & \textbf{81.73}\rlap{\impr{+45.67}} & $--$ & \textbf{78.28}\rlap{\impr{+44.02}} & $--$ & \textbf{50.00}\rlap{\impr{+39.82}} & $--$ & \textbf{74.89}\rlap{\impr{+20.99}} & $--$ & \textbf{83.45}\rlap{\impr{+40.06}} & $--$ \\
      & \textbf{STaR-8B}   & \cmark & \textbf{92.27}\rlap{\impr{+8.98}}  & $--$ & \textbf{92.96}\rlap{\impr{+22.92}} & $--$ & \textbf{56.06}\rlap{\impr{+29.43}} & $--$ & \textbf{97.36}\rlap{\impr{+32.60}} & $--$ & \textbf{92.05}\rlap{\impr{+1.83}}  & $--$ \\
    \bottomrule
  \end{tabular*}
\end{table*}

Table \ref{tab:overall-performance} summarizes the exact match (EM) results on all evaluated benchmarks, showing that STaR-8B achieves state-of-the-art performance. STaR-8B outperforms all baselines, including specialized table reasoning models, Table-R1 variants, and substantially larger general LLMs. The results also follow expected trends: models with slow-thinking capabilities outperform those without, larger models generally outperform smaller ones, and table-specialized systems outperform general LLMs at similar scale. Across all benchmarks, we report EM accuracy on the test splits under the same inference-time rollout setting (8 samples per query), enabling a consistent comparison of pass@1 performance after trajectory selection. 

The overall results are consistent with the contribution of each component in our training pipeline. SFT on self-verified demonstrations provides a strong initialization, as shown in Table \ref{tab:sft-rl-ablation}, while the complete framework integrating difficulty-aware RL and trajectory-level uncertainty quantification yields substantial gains: STaR-8B reaches 92.27\% on WTQ and 92.96\% on HiTab, representing +8.98 and +22.92 points over the base model. These gains support both our difficulty-aware design that guides models from simple to complex patterns and our selection mechanism of better trajectory, as analyzed in Table \ref{tab:answer-selection-comparison}. Notably, STaR also demonstrates strong out-of-domain performance on TabFact (92.05\%) and TabMWP (97.36\%), suggesting improved cross-task transfer beyond dataset-specific pattern memorization.

Despite its small size, STaR-0.6B achieves competitive performance compared to much larger baselines on several benchmarks, highlighting the parameter efficiency of our approach. Overall, STaR strikes an advantageous balance between performance and computational efficiency.

\subsection{Training Pipeline Analysis}

\begin{figure}[t]
  \centering
  \includegraphics[width=\linewidth]{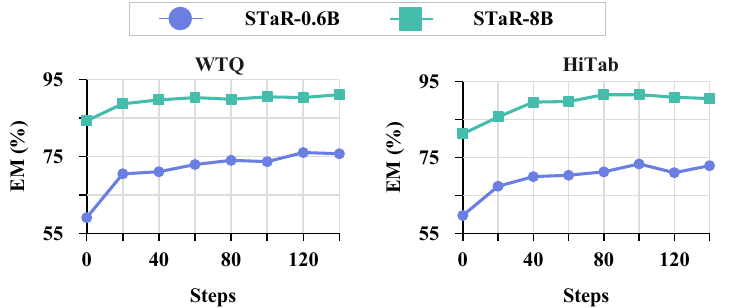}
  \caption{Training curves of STaR-0.6B and STaR-8B with difficulty-aware RL on WTQ and HiTab.}
  \label{fig:convergence-curve}
  \vspace{-0.5em}
\end{figure}

Figure \ref{fig:convergence-curve} plots the first 140 steps on WTQ and HiTab for difficulty-aware RL training of both model sizes. Both STaR-0.6B and STaR-8B improve: foundational training (first 20 steps) yields rapid gains on easy samples, and progressive training (next 120 steps) further improves reasoning abilities on harder samples. Foundational training uses a higher learning rate with fewer rollouts; progressive training uses a lower learning rate with more rollouts and dynamic filtering to focus updates on unsolved hard samples. The full difficulty-aware RL runs for 188 steps.

Unless otherwise noted, we conduct the following analyses on STaR-0.6B, where differences across settings are more clearly observed. To investigate the individual contributions of SFT and RFT, we conduct ablation experiments as presented in Table \ref{tab:sft-rl-ablation}. The results reveal complementary strengths: SFT-only training provides stable baseline performance across datasets but lacks the reasoning depth for complex queries, while RFT-only training achieves strong performance on some datasets (e.g., WTQ) but suffers from inconsistency, particularly on TabFact where performance drops dramatically. The complete SFT+RFT pipeline achieves the best performance across all benchmarks, demonstrating that SFT establishes reasoning foundations while RFT refines and deepens these capabilities. This synergy is particularly evident in FinQA and HiTab, where the combined approach substantially outperforms either component alone.

\subsection{RFT Components Ablation}

\begin{table}[t]
  \centering
  \caption{Ablation study on training components. Performance comparison of STaR-0.6B under different training configurations across benchmarks.}
  \label{tab:sft-rl-ablation}
  \small 
  \setlength{\tabcolsep}{3.5pt}
  \renewcommand{\arraystretch}{1.08}
  \begin{adjustbox}{width=\linewidth} 
    \begin{tabular}{@{}lccccc@{}}
      \toprule
      \textbf{Model Configuration} & \makecell{\textbf{WTQ}} & \textbf{HiTab} & \textbf{FinQA} & \textbf{TabMWP} & \textbf{TabFact}\\
      \midrule
      w/o RFT (SFT only) & 58.70 & 60.64 & 30.20 & 54.27 & 72.55 \\
      w/o SFT (RFT only) & 71.18 & 57.82 & 38.80 & 66.24 & 37.44 \\
      SFT + RFT (Full)   & 76.45 & 74.74 & 46.11 & 68.10 & 81.28 \\
      \bottomrule
    \end{tabular}
  \end{adjustbox}
\end{table}

\begin{table}[t]
  \centering
  \caption{Ablation study on difficulty-aware RL components for STaR-0.6B. All settings are trained from the SFT model.}
  \label{tab:two-stage-grpo-ablation}
  \small 
  \setlength{\tabcolsep}{3.5pt}
  \renewcommand{\arraystretch}{1.08}
  \begin{adjustbox}{width=\linewidth} 
    \begin{tabular}{@{}lccccc@{}}
      \toprule
      \textbf{Model Configuration} & \makecell{\textbf{WTQ}} & \textbf{HiTab} & \textbf{FinQA} & \textbf{TabMWP} & \textbf{TabFact}\\
      \midrule
      w/o phase 1  & 75.62 & 74.04 & 45.61 & 66.86 & 80.44 \\
      w/o phase 2   & 69.34 & 60.40 & 37.14 & 65.66 & 75.79 \\
      phase 1 + phase 2   & 76.45 & 74.74 & 46.11 & 68.10 & 81.28 \\
      \bottomrule
    \end{tabular}
  \end{adjustbox}
\end{table}

\begin{table}[t]
  \centering
  \caption{Impact of uncertainty quantification (UQ) on model performance. Comparison of STaR-0.6B with and without selection mechanism.}
  \label{tab:answer-selection-comparison}
  \begin{adjustbox}{width=\linewidth} 
    \begin{tabular}{@{}lccccc@{}}
      \toprule
      \textbf{Model Configuration} & \makecell{\textbf{WTQ}} & \textbf{HiTab} & \textbf{FinQA} & \textbf{TabMWP} & \textbf{TabFact}\\
      \midrule
      w/o UQ            & 76.45 & 74.75 & 46.11 & 68.10 & 82.28 \\
      STaR (with UQ)    & 81.73 & 78.28 & 50.00 & 74.89 & 83.45 \\
      \bottomrule
    \end{tabular}
  \end{adjustbox}
\end{table}

\begin{figure}[t]
  \centering
  \includegraphics[width=\linewidth]{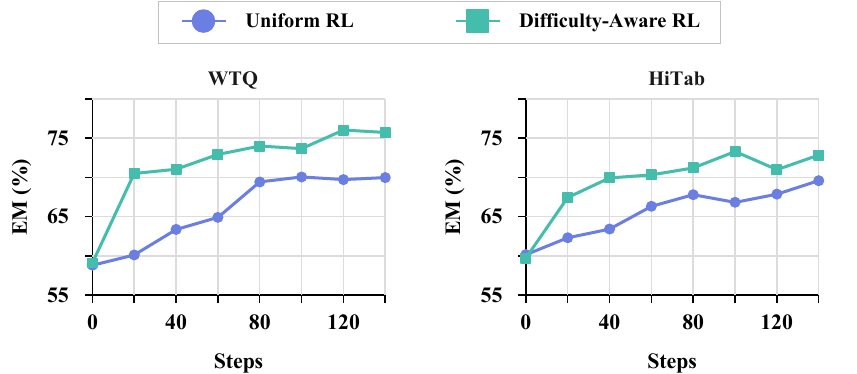}
  \caption{Comparison of uniform and difficulty-aware RL for STaR-0.6B on WTQ and HiTab.}
  \label{fig:grpo-comparation}
\end{figure}

Our difficulty-aware RL in the RFT stage comprises two phases: phase~1 (foundational training on easy samples) and phase~2 (progressive training on hard samples). To assess the effect of this design, we compare uniform versus difficulty-aware training curves for STaR-0.6B in Figure \ref{fig:grpo-comparation}, which plots the first 140 steps on WTQ and HiTab. The difficulty-aware approach yields faster early gains: by step 20, phase~1 enables the model to surpass the uniform baseline, which improves more gradually. On WTQ, the uniform baseline stabilizes around step 120, whereas the difficulty-aware method continues improving. Overall, these trends suggest that separating easy and hard samples into distinct training phases accelerates initial learning while preserving room for further refinement.

To further investigate the contribution of each phase, we conduct ablation experiments presented in Table \ref{tab:two-stage-grpo-ablation}. Removing phase~2 reduces performance across all datasets, with drops of 14.34 points on HiTab and 8.97 points on FinQA, suggesting that focused training on hard samples is important for achieving strong final performance. While removing phase~1 leads to smaller final performance differences, Figure \ref{fig:grpo-comparation} suggests its role in training efficiency: rapid learning on easy samples during phase~1 provides a strong starting point that accelerates overall training. Overall, the complete difficulty-aware RL pipeline achieves the best results by combining efficient bootstrapping via phase~1 with targeted refinement via phase~2, consistent with our difficulty-aware design motivation.

\begin{figure}[t]
  \centering
  \includegraphics[width=\linewidth]{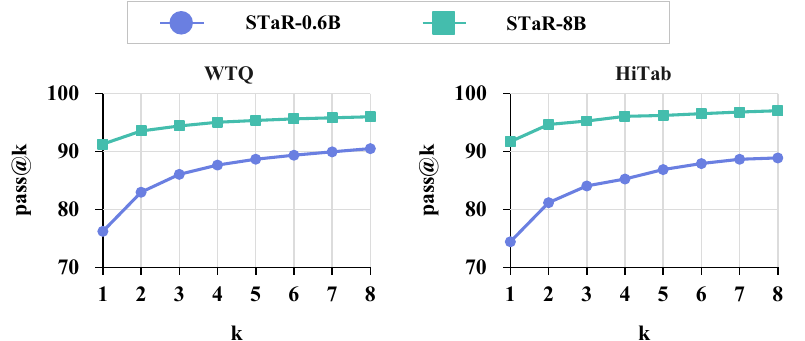}
  \caption{Pass@k accuracy curves for STaR models on WTQ and HiTab benchmarks.}
  \label{fig:pass@k-curve}
\end{figure}

\subsection{Impact of Uncertainty Quantification}

\begin{figure*}[t]
  \centering
  \includegraphics[width=\textwidth]{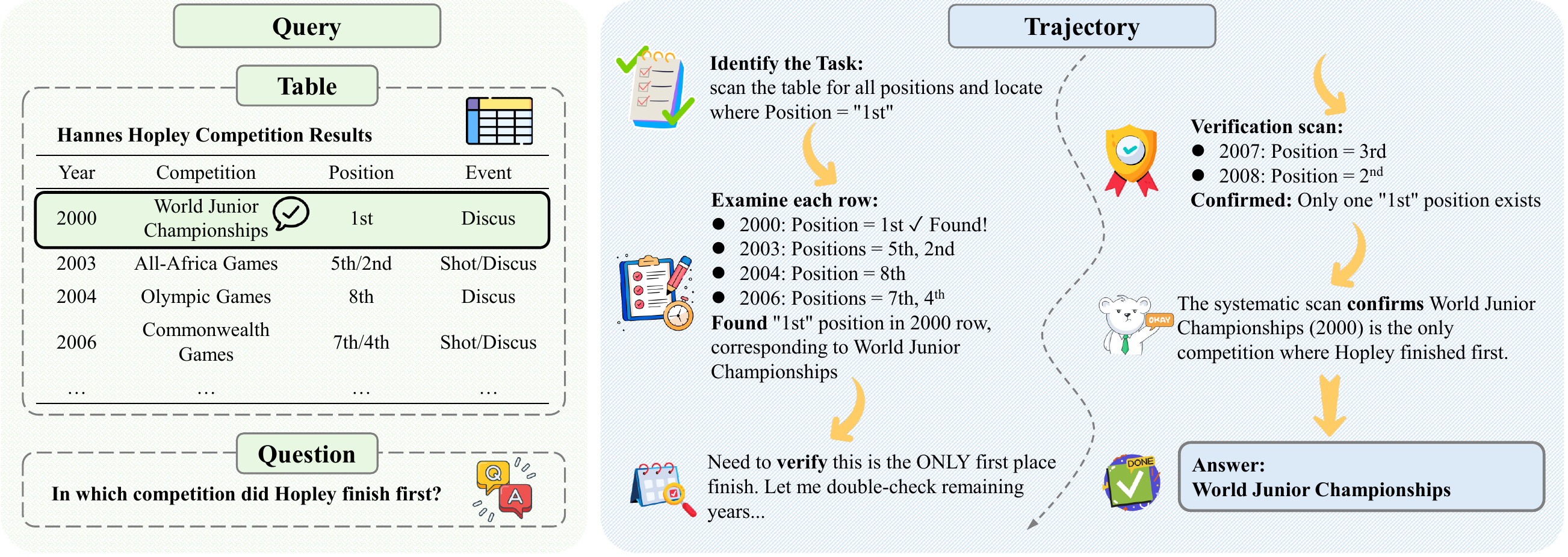}
  \caption{A case study demonstrates STaR's slow-thinking reasoning process.}
  \label{fig:case-study}
\end{figure*}

To analyze our uncertainty quantification (UQ) inference approach, we report pass@k accuracy curves as shown in Figure \ref{fig:pass@k-curve}, where pass@k counts a query as correct if any of the $k$ sampled rollouts matches the ground truth. Both STaR-0.6B and STaR-8B exhibit higher pass@k as $k$ increases; on WTQ, pass@8 reaches approximately 90\% and 96\% respectively, corresponding to gains of about 14 and 5 points over pass@1. The gap between pass@1 and pass@k suggests that correct reasoning trajectories are often present among sampled candidates but are not always selected as the final answer. Our UQ-based selection is designed to convert this latent pass@k potential into higher pass@1 accuracy, as reflected in Table \ref{tab:answer-selection-comparison}.

To quantify the contribution of our UQ mechanism, we compare model performance with and without UQ in Table \ref{tab:answer-selection-comparison}. The model with UQ improves on all benchmarks, with the largest gains on TabMWP (+6.79 points) and WTQ (+5.28 points). These results suggest that our trajectory-level UQ, which fuses token-level confidence with answer-level consistency, helps select better reasoning trajectories from multiple candidates, improving pass@1 accuracy. The gains are larger on datasets requiring complex multi-step reasoning, where distinguishing between plausible but incorrect trajectories and truly valid solutions is important.

\begin{figure}[t]
  \centering
  \includegraphics[width=\linewidth]{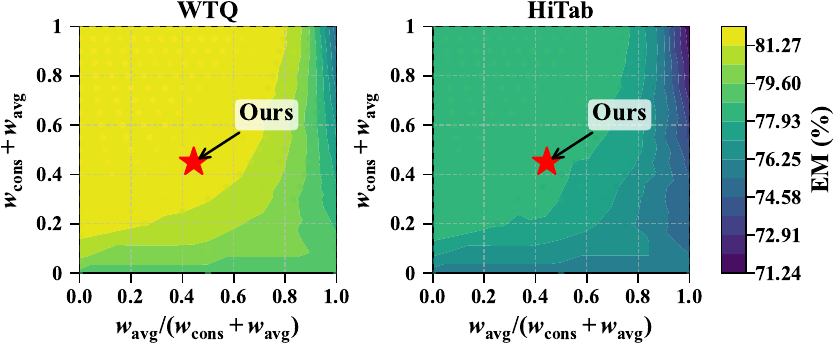}
  \caption{Uncertainty quantification fusion-weight sensitivity on WTQ and HiTab for STaR-0.6B. The red star marks our setting (0.25, 0.20, 0.55).}
  \label{fig:uq-weight-sensitivity}
  \vspace{-1em}
\end{figure}

\subsection{Sensitivity Analysis}
We assess the sensitivity of inference-time trajectory selection to the uncertainty quantification fusion weights in Algorithm~\ref{alg:ccfs-2e}. Figure \ref{fig:uq-weight-sensitivity} visualizes the resulting EM landscape on WTQ and HiTab for STaR-0.6B. We sweep the weight triplet \((w_{\text{cons}}, w_{\text{avg}}, w_{\text{max}})\) on the simplex \(w_{\text{cons}}+w_{\text{avg}}+w_{\text{max}}=1\) and report EM on the WTQ and HiTab test splits for STaR-0.6B. To visualize the simplex without leaving an empty triangular region, we map it to a unit square with \(y=w_{\text{cons}}+w_{\text{avg}}(=1-w_{\text{max}})\) and \(x=\frac{w_{\text{avg}}}{w_{\text{cons}}+w_{\text{avg}}}\). Under this parameterization, \(y\) controls the total weight assigned to \((w_{\text{cons}}+w_{\text{avg}})\) versus \(w_{\text{max}}\), while \(x\) controls the split between \(w_{\text{cons}}\) and \(w_{\text{avg}}\) when \(y>0\). The mapping is invertible: \(w_{\text{avg}}=xy\), \(w_{\text{cons}}=(1-x)y\), and \(w_{\text{max}}=1-y\). We discretize the simplex with a step size of 0.05; dots indicate evaluated settings and the background is an interpolation for readability. The broad high-performing region suggests that our selection mechanism is stable under moderate changes in the fusion weights. Our setting \((w_{\text{cons}}, w_{\text{avg}}, w_{\text{max}})=(0.25, 0.20, 0.55)\) (red star), selected via grid search on the training splits, stays within only 0.5 EM points of the best setting in the sweep on both datasets.

Intuitively, \(x\) trades off \(w_{\text{avg}}\) against \(w_{\text{cons}}\) given \(y\), while \(y\) trades off \((w_{\text{cons}}+w_{\text{avg}})\) against \(w_{\text{max}}\). Performance is relatively flat for moderate allocations and degrades at extremes. The red star need not coincide with the center of the high-performing region because the landscape is anisotropic and differs across datasets, and we use a single setting selected on training splits. Being close to a contour reflects a direction of larger change rather than fragility.

\subsection{Case Study Visualization}
Figure \ref{fig:case-study} provides an illustrative example of STaR's slow-thinking reasoning trajectory on a representative table question answering task. When asked "In which competition did Hopley finish first?", the model generates an explicit multi-step reasoning trajectory instead of answering immediately. The thinking process begins with explicit task identification, where the model recognizes the need to scan for Position = "1st" across all table entries. STaR reviews the rows from earliest to latest and locates the first place finish at the 2000 World Junior Championships. Notably, rather than stopping at the first match, the model performs a final verification pass by continuing to examine the remaining years (2007–2008), confirming that this represents the only instance of a first place finish and thereby avoiding premature or incomplete conclusions. By explicitly exposing the intermediate steps, the reasoning trajectory becomes easier to inspect and debug. Each intermediate decision is grounded in concrete table entries, which helps readers verify how the final answer is derived. This example also shows how a verification step, where the model explicitly examines other relevant table entries for confirmation, can be reflected in the reasoning trace. We include this case study as a qualitative complement to the quantitative evaluations above.

\section{Conclusion}
In this paper, we proposed STaR, a slow-thinking model for effective and stable table reasoning. To address the limitations of insufficient reasoning depth and instability in existing approaches, we design a two-stage training framework consisting of SFT warm-up with self-verified demonstrations followed by RFT with difficulty-aware RL. Furthermore, to improve reasoning stability, we introduce trajectory-level uncertainty quantification that fuses token-level confidence with answer-level consistency to select better reasoning trajectories. Experiments demonstrate that STaR-8B achieves state-of-the-art performance on in-domain benchmarks and exhibits strong generalization to out-of-domain datasets, highlighting its potential for enhancing both effectiveness and stability in table reasoning. Future work could extend STaR to multi-table scenarios, incorporate visual table understanding, and explore the applicability of our approach to broader tabular reasoning domains.

\begin{acks}
This research was supported by grants from the National Natural Science Foundation of China (No. 62502486), the Provincial Natural Science Foundation of Anhui Province (No. 2408085QF193), USTC Research Funds of the Double First-Class Initiative (No. YD2150002501), and the Fundamental Research Funds for the Central Universities of China (No. WK2150110032).
\end{acks}

\bibliographystyle{ACM-Reference-Format}
\bibliography{main}

\appendix

\section{Method Details}

\subsection{Difficulty-Aware RL Pipeline}
\label{sec:difficulty-aware-rl-pipeline}

\begin{figure}[t]
  \centering
  \includegraphics[width=\linewidth]{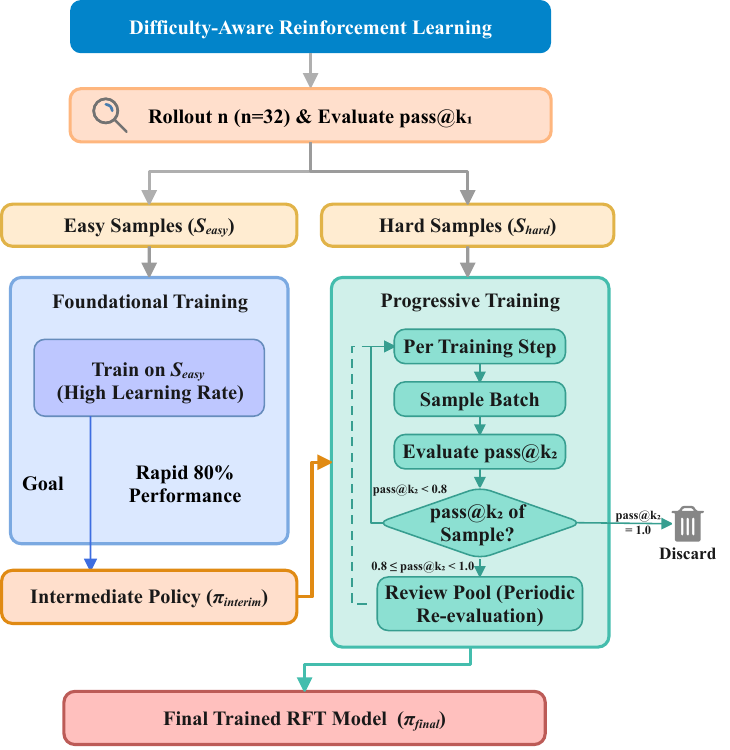}
  \caption{Difficulty-aware RL pipeline with dataset partitioning and dynamic sample filtering.}
  \label{fig:two-stage-grpo}
  \vspace{-1em}
\end{figure}

Figure \ref{fig:two-stage-grpo} illustrates our difficulty-aware RL pipeline. We split training samples into an easy subset $S_{\text{easy}}$ and a hard subset $S_{\text{hard}}$ using pass@k$_1$ with $k_1{=}n{=}32$ rollouts and an SFT-estimated threshold of 0.6, yielding approximately 10,000 samples for each subset. Here, pass@k counts a query as correct if any of the $k$ sampled trajectories matches the ground truth \cite{chen2021evaluating}.

\noindent \textbf{Foundational Training.} Training on $S_{\text{easy}}$ with high learning rates (e.g., $1\times 10^{-5}$) produces the intermediate policy $\pi_{\text{interim}}$. This stage rapidly achieves approximately 80\% performance within minimal training steps by focusing on samples the model can already handle reasonably well.

\noindent \textbf{Progressive Training.} Starting from $\pi_{\text{interim}}$, we train on $S_{\text{hard}}$ with lower learning rates (e.g., $1\times 10^{-6}$) and apply dynamic sample filtering based on pass@k$_2$ evaluation. As shown in the figure, samples with pass@k$_2$=1.0 are permanently discarded; samples with $0.8 \le \text{pass@k}_2 < 1.0$ enter a review pool for periodic re-evaluation; only samples with pass@k$_2$<0.8 receive active GRPO updates. This filtering mechanism ensures computational resources focus on challenging problems at the appropriate difficulty level.

\section{Experimental Details}

\begin{table*}[t]
  \centering
  \caption{Full overall performance comparison on in-domain (WTQ, HiTab, FinQA) and out-of-domain (TabMWP, TabFact) datasets. Metric: EM accuracy (\%).}
  \label{tab:full-overall-performance}
  \small
  \setlength{\tabcolsep}{2.5pt}
  \renewcommand{\arraystretch}{1.12}
  \sisetup{table-number-alignment = center, detect-weight = true, detect-inline-weight = math}
  \begin{tabular*}{\textwidth}{@{\extracolsep{\fill}} l l l
      S[table-format=3.2] S[table-format=3.2] S[table-format=3.2]
      S[table-format=3.2] S[table-format=3.2] @{}}
    \toprule
    \multirow{2}{*}{\textbf{Type}} & \multirow{2}{*}{\textbf{Subtype}} & \multirow{2}{*}{\textbf{Model}} &
      \multicolumn{3}{c}{\textbf{In-domain Performance}} &
      \multicolumn{2}{c}{\textbf{Out-of-domain Performance}} \\
    \cmidrule(lr){4-6}\cmidrule(lr){7-8}
     &  &  & \multicolumn{1}{c}{\textbf{WTQ}} & \multicolumn{1}{c}{\textbf{HiTab}} & \multicolumn{1}{c}{\textbf{FinQA}}
                 & \multicolumn{1}{c}{\textbf{TabMWP}} & \multicolumn{1}{c}{\textbf{TabFact}} \\
    \midrule
    \multirow{4}{*}{Closed-Source} 
      & \multirow{2}{*}{Non-Thinking} & GPT-4.1 & 58.42 & 60.64 &  6.36 & 37.73 & 84.80 \\
      &                                & Gemini-2.0-flash & 71.81 & 76.95 & 19.35 & 61.96 & 80.25 \\[2pt]
      & \multirow{2}{*}{Slow-Thinking} & GPT-5 & 90.10 & 43.96 & 29.21 & 55.18 & 91.20 \\
      &                                   & Gemini-2.5-flash & 82.12 & 52.11 & 15.69 & 62.21 & 90.35 \\[2pt]
    \specialrule{0.08em}{0pt}{2pt}
    \multirow{7}{*}{Open-Source} 
      & \multirow{3}{*}{Non-Thinking} & Qwen2.5-7B-Instruct  & 49.74 & 62.48 & 16.58 & 55.02 & 45.54 \\
      &                                & Qwen2.5-72B-Instruct & 67.56 & 71.46 & 29.47 & 74.33 & 77.90 \\
      &                                & DeepSeek V3          & 68.55 & 75.39 & 21.45 & 65.10 & 78.35 \\[2pt]
      & \multirow{4}{*}{Slow-Thinking} & Qwen3-8B       & 83.29 & 70.04 & 26.63 & 64.76 & 90.22 \\
      &                                   & Qwen3-235b-a22b & 83.97 & 80.73 & 32.61 & 82.23 & 89.35 \\
      &                                   & DeepSeek V3.1   & 84.07 & 65.68 & 24.93 & 71.06 & 89.50 \\
      &                                   & GPT-oss-120b    & 81.18 & 41.66 & 17.70 & 61.90 & 89.45 \\[2pt]
    \specialrule{0.08em}{0pt}{2pt}
    \multirow{5}{*}{\makecell[l]{Table-Reasoning\\Specific}} 
      & TableGPT Series                 & TableGPT2-7B         & 47.60 & 63.11 & 15.54 & 53.71 & 21.16 \\[2pt]
      & \multirow{4}{*}{Table-R1 Series} & Table-R1-SFT-8B   & 84.24 & 85.95 & 14.61 & 54.73 & 90.25 \\
      &                                   & Table-R1-Zero-8B  & 83.09 & 88.61 & 14.46 & 43.78 & 87.63 \\
      &                                   & Table-R1-SFT-7B   & 81.55 & 81.23 & 21.04 & 66.67 & 88.76 \\
      &                                   & Table-R1-Zero-7B  & 82.04 & 86.63 & 19.23 & 71.50 & 56.52 \\[2pt]
    \specialrule{0.08em}{0pt}{2pt}
    \multirow{6}{*}{Our Base and Ours} 
      & \multirow{3}{*}{0.6B Models} & Qwen3-0.6B                & 36.06 & 34.26 & 10.18 & 53.90 & 43.39 \\
      &                               & Qwen3-0.6B-SFT            & 58.70 & 60.64 & 30.20 & 54.27 & 72.55 \\
      &                               & STaR-0.6B                 & 81.73 & 78.28 & 50.00 & 74.89 & 83.45 \\[2pt]
      & \multirow{3}{*}{8B Models}   & Qwen3-8B                  & 83.29 & 70.04 & 26.63 & 64.76 & 90.22 \\
      &                               & Qwen3-8B-SFT              & 82.33 & 80.74 & 42.60 & 80.70 & 88.57 \\
      &                               & STaR-8B                   & 92.27 & 92.96 & 56.06 & 97.36 & 92.05 \\[2pt]
    \bottomrule
  \end{tabular*}
\end{table*}

\subsection{Dataset Statistics}
\label{sec:dataset-statistics}

Table \ref{tab:dataset-statistics} presents detailed dataset statistics. We employ three in-domain datasets for training and evaluation: WikiTableQuestions (WTQ), HiTab, and FinQA, constructing high-quality slow-thinking demonstrations using the self-verification mechanism from section \ref{sec:sft}. We use the standard train/test splits provided by each dataset. The training splits are used for both SFT and RFT, where pass@k-based difficulty assessment partitions training samples into easy and hard subsets for difficulty-aware RL. The test splits are used only for final evaluation. For out-of-domain generalization, we evaluate on TabMWP and TabFact, used only for inference without training involvement.

\begin{table}[h]
  \centering
  \caption{Dataset statistics for training and evaluation. In-domain datasets (WTQ, HiTab, FinQA) are used for both training and evaluation, while out-of-domain datasets (TabMWP, TabFact) are used only for evaluation.}
  \label{tab:dataset-statistics}
  \small 
  \setlength{\tabcolsep}{3.5pt}
  \renewcommand{\arraystretch}{1.08}
  \begin{adjustbox}{width=\linewidth} 
      \begin{tabular}{@{}cccccc@{}}
        \toprule
        \textbf{Split} & \textbf{WTQ} & \textbf{HiTab} & \textbf{FinQA} & \textbf{TabMWP} & \textbf{TabFact}\\
        \midrule
        Training   & 12,263 & 5,665 & 6,251 & -- & -- \\
        Test       & 3,937  & 1,349 & 1,138 & 1,593 & 2,000 \\
        \bottomrule
      \end{tabular}
  \end{adjustbox}
\end{table}

\noindent \textbf{In-domain datasets:} WTQ, HiTab, and FinQA cover diverse table reasoning scenarios including open-domain question answering, hierarchical table reasoning, and financial numerical reasoning. Training samples are partitioned via pass@k difficulty assessment, while test splits are reserved for final evaluation.

\noindent \textbf{Out-of-domain datasets:} TabMWP focuses on mathematical word problems grounded in tables, while TabFact requires binary fact verification. These datasets evaluate the model's ability to generalize reasoning capabilities to unseen task formats without any direct training data exposure.

\subsection{Prompt Templates}
\label{sec:prompt-templates}

To ensure reproducibility, we provide the exact prompt templates and table serialization format used in our experiments.

\subsubsection{SFT Prompt (Single Message)}
The SFT data uses a single user prompt without a system prompt:
{\footnotesize
\begin{verbatim}
Instruction
Answer the question based on the provided table.

Table
Table Title: [TABLE_TITLE]
Table Content:
[TABLE_CONTENT]

Question
[QUESTION]

Answer Format
The final answer should be concise and use the following format:
{
  "answer": [
    "answer1",
    "answer2",
    ...
  ]
}
\end{verbatim}
}

\subsubsection{GRPO Prompt (Chat-Style)}
Both difficulty-aware RL phases use chat-style prompts with separate system and user messages. The system prompt is:
{\footnotesize
\begin{verbatim}
A conversation between User and Assistant. The user asks a question,
and the assistant solves it.
The assistant first thinks about the reasoning process in the mind
and then provides the user with the answer.
The reasoning process and answer are enclosed within <think> </think>
and <answer> </answer> tags, respectively, i.e.,
<think> reasoning process here </think>
<answer> answer here </answer>.
\end{verbatim}
}
The user prompt is identical to the SFT prompt above.

\subsubsection{Table Serialization}
In all experiments, \texttt{[TABLE\_CONTENT]} is instantiated as a plain-text markdown table string using pipe (\texttt{|}) syntax with one header row, one alignment row, and one row per record. Column and row order are preserved from the original dataset, and cell values are inserted as-is.

\subsection{Full Overall Performance Results}
\label{sec:full-overall-performance}

Table \ref{tab:full-overall-performance} provides a comprehensive comparison of STaR against baseline categories across all evaluation benchmarks. The baselines encompass: (1) closed-source models without slow-thinking capabilities (GPT-4.1, Gemini-2.0-flash), (2) closed-source models with slow-thinking capabilities (GPT-5, Gemini-2.5-flash), (3) open-source general LLMs without slow-thinking capabilities (Qwen2.5-7B/72B-Instruct, DeepSeek V3), (4) open-source models with slow-thinking capabilities (Qwen3-8B/235B, DeepSeek V3.1, GPT-oss-120B), and (5) specialized table reasoning systems (TableGPT2-7B, Table-R1 variants with both SFT and RFT).

\begin{figure}[t]
  \centering
  \includegraphics[width=\linewidth]{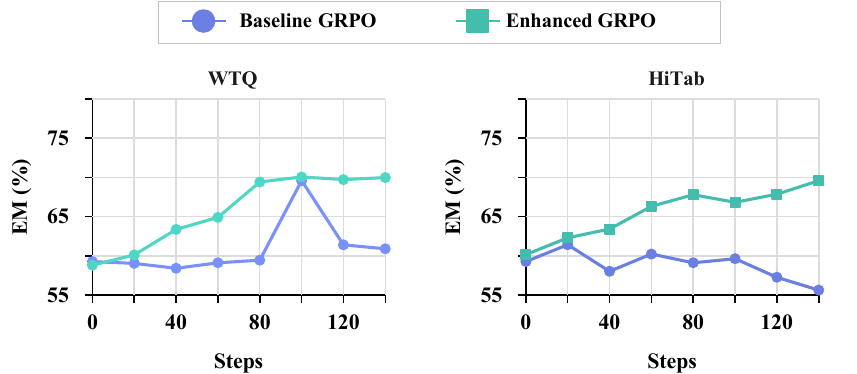}
  \caption{Comparison of Enhanced GRPO versus Original GRPO on WTQ and HiTab benchmarks.}
  \label{fig:enhance-grpo-comparation}
\end{figure}

\subsection{Enhanced GRPO and Baseline Comparison}
\label{sec:enhance-grpo-comparation}

\noindent We compare the enhanced GRPO variant adopted in STaR (i.e., the DAPO formulation without KL penalty and with asymmetric clipping) with the original GRPO objective under the same training setup on WTQ and HiTab training splits, as shown in Figure \ref{fig:enhance-grpo-comparation}. The enhanced GRPO variant shows smoother and more consistent improvements, while the original GRPO exhibits larger oscillations and occasional degradation at later steps. These observations are consistent with the motivation for adopting the DAPO-style GRPO objective to stabilize RL optimization for table reasoning.


\end{document}